\def\year{2022}\relax
\documentclass[letterpaper]{article} 
\usepackage{aaai22}  
\usepackage{times}  
\usepackage{helvet}  
\usepackage{courier}  
\usepackage[hyphens]{url}  
\usepackage{graphicx} 
\usepackage{natbib}  
\usepackage{caption} 
\DeclareCaptionStyle{ruled}{labelfont=normalfont,labelsep=colon,strut=off} 
\usepackage{algorithm}
\usepackage{algorithmic}
\usepackage{amsmath} 
\usepackage{amsfonts,amssymb}

\usepackage{newfloat}
\usepackage{listings}
\floatstyle{ruled}
\newfloat{listing}{tb}{lst}{}
\floatname{listing}{Listing}
\usepackage{adjustbox}
\usepackage{latexsym}
\usepackage{todonotes}
\usepackage{array}
\usepackage{subfigure}

\title{StepGame: A New Benchmark for \\ Robust Multi-Hop Spatial Reasoning in Texts}
\author {
    Zhengxiang Shi\textsuperscript{\rm 1},
    Qiang Zhang\textsuperscript{\rm 2},
    Aldo Lipani\textsuperscript{\rm 1}
}
\affiliations {
    \textsuperscript{\rm 1}University College London\\
    \textsuperscript{\rm 2}Zhejiang University\\
    zhengxiang.shi.19@ucl.ac.uk, qiang.zhang.cs@zju.edu.cn, aldo.lipani@ucl.ac.uk
}

\usepackage{bibentry}

\begin{document}
\maketitle

\begin{abstract}
Inferring spatial relations in natural language is a crucial ability an intelligent system should possess. The bAbI dataset tries to capture tasks relevant to this domain (task 17 and 19). However, these tasks have several limitations. Most importantly, they are limited to fixed expressions, they are limited in the number of reasoning steps required to solve them, and they fail to test the robustness of models to input that contains irrelevant or redundant information. In this paper, we present a new Question-Answering dataset called StepGame for  \textit{robust multi-hop spatial reasoning in texts}. Our experiments demonstrate that state-of-the-art models on the bAbI dataset struggle on the StepGame dataset. Moreover, we propose a Tensor-Product based Memory-Augmented Neural Network (TP-MANN) specialized for spatial reasoning tasks. Experimental results on both datasets show that our model outperforms all the baselines with superior generalization and robustness performance.

\end{abstract}

\section{Introduction}
Neural networks have been successful in a wide array of perceptual tasks, but it is often stated that they are incapable of solving tasks that require higher-level reasoning~\cite{ding2020object}. %
Since spatial reasoning is ubiquitous in many scenarios such as %
autonomous navigation~\cite{vogel2010learning}, %
situated dialog~\cite{kruijff2007situated}, and %
robotic manipulation~\cite{yang2020robust, landsiedel2017review},
grounding spatial references in texts is essential for effective human-machine communication through natural language. 
Navigation tasks require agents to reason about their relative position to objects and how these relations change as they move through the environment~\cite{chen2019touchdown}. 
If we want to develop conversational systems able to assist users in solving tasks where spatial references are involved, we need to make them able to understand and reason about spatial references in natural language.
Such ability can help conversational systems to successfully follow instructions and understand spatial descriptions. However,
despite its tremendous applicability, reasoning over spatial relations 
remains a challenging task for existing conversational systems.

Earlier works in spatial reasoning focused on spatial instruction understanding in a synthetic environment~\cite{bisk2018learning, tan2018source, janner2018representation} or in a simulated world with spatial information annotation in texts~\cite{pustejovsky2015semeval}, spatial relation extractions across entities~\cite{petruck2018representing} and visual observations~\cite{anderson2018vision, chen2019touchdown}. 
However, few of the existing datasets are designed to evaluate models' inference over spatial information in texts. A spatial relational inference task often requires an conversational system to infer the spatial relation between two items given a description of a scene. For example, imagine a user asking to a conversational system to recognize the location of an entity based on the description of other entities in a scene. To do so, the conversational system needs to be able to reason about the location of the various entities in the scene using only textual information.

BAbI~\cite{weston2015towards} is the most relevant dataset for this task. It contains 20 synthetic question answering (QA) tasks to test a variety of reasoning abilities in texts, 
like deduction, co-reference, and counting. 
In particular, the \textit{positional reasoning} task (no.~17) and the \textit{path finding} task (no.~19) are designed to evaluate models' spatial reasoning ability.
These two tasks are arguably the most challenging ones~\citet{van2019does}. 
The state-of-the-art model on the bAbI~\cite{le2020self} dataset almost perfectly solve these two spatial reasoning tasks. 
However, in this paper, we demonstrate that such good performance is attributable to issues with the bAbI dataset rather than the model inference ability.

We find four major issues with bAbI's tasks 17 and 19: 
(1) 
There is a data leakage between the train and test sets; that is, most of the test set samples appear in the training set. Hence, the evaluation results on the test set cannot truly reflect models' reasoning ability; 
(2) 
Named entities are fixed and only four relations are considered. Each text sample always contains the same four named entities in the training, validation, and test sets. This further biases the learning models towards these four entities. When named entities in the test set are replaced by unseen entities or the number of such entities increases, the model performance decreases dramatically~\cite{chen2020unseen}. Also, relations such as top-left, top-right, lower-left, lower-right are not taken into consideration;
(3) 
Learning models are required to reason only over one or two sentences in the text descriptions, making such tasks relatively simple. \citet{palm2018recurrent} pointed out that multi-hop reasoning is not necessary for the bAbI dataset since models only need a single step to solve all the tasks, and;
(4) It is a synthetic dataset with a limited diversity of spatial relation descriptions. It thus cannot truly reveal the models' ability in understanding textual space descriptions.

In this paper, we propose a new dataset called \textit{StepGame} to tackle the above-mentioned issues and a novel \textit{Tensor Product-based Memory-Augmented Neural Network} architecture (TP-MANN) for multi-hop spatial reasoning in texts.  

The StepGame dataset is based on crowdsourced descriptions of 
8 potential spatial relations between 2 entities.
These descriptions are then used as templates when generating the dataset. 
To increase the diversity of these templates, crowdworkers were asked to diversify their expressions. 
This was done in order to ensure that the crowdsourced templates cover most of the natural ways relations between two entities can be described in text.
The StepGame dataset is characterized by a combinatorial growth in the number of possible description of scenes, named \textit{stories}, as the number of described relations between two entities increases. 
This combinatorial growth reduces the chances to leak stories from the training to the validation and test sets. 
Moreover, we use a large number of named entities and 
require multi-hop reasoning to answer \textit{questions} about two entities mentioned in the stories. %
Experimental results show that existing models (1) fail to achieve a performance on the StepGame dataset similar to that achieved on the bAbI dataset, and (2) suffer from a large performance drop as the number of required reasoning steps increases.

The TP-MANN architecture is based on tensor product representations~\cite{smolensky1990tensor} that are used in a recurrent memory module to store, update or delete the relation information among entities inferred from stories. 
This recurrent architecture provides three key benefits: 
(1) it enables the model to make inferences based on the stored memory; (2) it allows multi-hop reasoning and it is robust to noise, and; (3) the number of parameters remains unchanged as the number of recurrent layers in the memory module increases.
Experimental results on the StepGame dataset show that 
our model achieves state-of-the-art performance with a substantial improvement, and demonstrates a better generalization ability to more complex stories. 
Finally, we also conduct some analysis of our recurrent structure and demonstrate its importance for multi-hop reasoning.

\section{Related Work and Background}
\subsection{Related Work}

\paragraph{Reasoning Datasets.}
The role of language in spatial reasoning has been investigated since the 1980s~\cite{herskovits1987language, gershman2015phrase, tversky2019mind}, and
reasoning about spatial relations has been studied in several contexts such as, 2D and 3D navigation~\cite{bisk2018learning, tan2018source, janner2018representation,yang2020robust}, and robotic manipulation~\cite{landsiedel2017review}. However, few of the datasets used in these works are used to evaluate systems' spatial reasoning ability in texts.

The bAbI~\cite{weston2015towards} dataset consists of several QA tasks. Solving these tasks require logical reasoning steps and cannot be solved by simply word matching. Of particular interest to this paper are tasks 17 and 19. Task 17 is about positional reasoning while task 19 is about path finding. These two tasks can be used to evaluate the spatial inference ability of learning models. However, the bAbI dataset has several issues as mentioned above: the data leakage, the fixed named entities and expressions, and the lack of a need to perform multi-hop reasoning. 
Another relevant dataset is SpartQA~\cite{mirzaee2021spartqa}, which is designed for spatial reasoning over texts but only requires a limited multi-hop reasoning compared to StepGame.

\paragraph{Multi-Hop QA Datasets.}
The multi-hop QA tasks require reasoning over multiple pieces of evidence and focus on leveraging the connections between entities to infer a requested property of a set of them. Commonly-used multi-hop QA datasets are HotpotQA~\cite{yang2018hotpotqa}, Complex\-Web\-Questions~\cite{talmor2018web}, and QAngaroo~\cite{welbl2018constructing}. The proposed StepGame dataset is different from these datasets.
The StepGame dataset focuses on spatial reasoning, which requires machine learning models to infer  
the spatial relations among the described entities. Moreover, multi-hop QA datasets usually require no more than two reasoning steps, while the StepGame dataset can require as many as 10 reasoning steps.

\paragraph{Reasoning Models.}
There are three types of reasoning models: 
memory-augmented neural networks, %
graph neural networks, and %
transformer-based networks. %
Works of the first type augment neural networks with external memory, such as End to End Memory Networks~\cite{sukhbaatar2015end}, Differential Neural Computer~\cite{graves2016hybrid}, and Gated End-to-End Memory Networks~\cite{liu2017gated}. These models have shown remarkable abilities in tackling difficult computational and reasoning tasks. 
Works of the second type use graph structure to 
incorporate a stronger relational inductive bias~\cite{battaglia2018relational}.
\citet{santoro2017simple} introduced Relational Networks (RN) and demonstrated strong relational reasoning capabilities with a shallow architecture by modelling binary relations between entity pairs. 
\citet{palm2018recurrent} proposed a graph representation of objects and models multi-hop relational reasoning using a message passing mechanism. 
Works of the third type use transformers. 
Although transformers have been proven successful in many NLP tasks, they still struggle with reasoning tasks. \citet{van2019does} analyzed the performance of BERT~\cite{devlin2018bert} on bAbI's tasks and demonstrated that most of BERT's errors 
come from task 17 and 19 which require spatial reasoning. Meanwhile, \citet{dehghani2018universal} demonstrated that standard transformers cannot perform as well as memory-augmented networks on the bAbI dataset.  
Moreover, it is important to note that most of the errors of their proposed Universal Transformer come also from task 17 and task 19 of the bAbI dataset, which matches our observations on other transformer-based models.
Therefore, spatial reasoning tasks are arguably the most challenging tasks in the bAbI dataset.

\paragraph{Tensor Product Representation.}
The Tensor  Product  Representation (TPR)~\cite{smolensky1990tensor, schlag2020learning} is a 
technique for encoding symbolic structural information and modelling symbolic reasoning in vector spaces by learning to deconstruct natural language statements into combinatorial representations~\cite{chen2020mapping}. 
TPR has been used for tasks that require deductive reasoning abilities and it is able to represent entire problem statements to solve math questions in natural language~\cite{chen2020mapping} and generate natural language captions from images~\cite{huang2018tensor}. %

\citet{schlag2018learning} proposed a gradient-based RNN with third-order TPR (TPR-RNN), which creates a vector space embedding of complex symbolic structures by tensor products and stores these learned representations into a third-order TPR-like memory. 
Self-Attentive Associative Memory (STM)~\cite{le2020self} utilizes a second-order item memory and a third-order TPR-like relational memory to simulate the hippocampus, achieving state-of-the-art performance on the bAbI dataset.
Despite a gain in the performance on bAbI compared to TPR-RNN, STM takes a longer time to converge in practice.

Recently, \citet{schlag2020learning} compared a concatenated memory $\boldsymbol{M} \in \mathbb{R}^{2d \times d}$ with a 3-order memory $\boldsymbol{M} \in \mathbb{R}^{d^{2} \times d}$, and experimental results indicate a drop in performance when a concatenated memory is used. 
However, neither STM nor TPR-RNN processes information at the paragraph level and allows later modifications after the first information is stored, 
as done in our model. 
Both STM and TPR-RNN use an RNN-like architecture where each sentence in a paragraph is stored recurrently. This may result in a long-term dependency problem~\cite{vaswani2017attention} where necessary information would not interact with each other. 
To solve this issue, an explicit mechanism to update relational information between entities at the end of each story
is introduced in our model.

\subsection{Background}

The Tensor Product Representation (TPR) is a method to create a vector space embedding of complex symbolic structures by tensor product. 
Such representation can be constructed as follows:
\begin{equation}
    \boldsymbol{M}=\sum_{i} f_i\otimes r_i= \sum_{i} f_i r_i^{^\top} = \sum_{i} (\boldsymbol{f} \otimes \boldsymbol{r})_{ii},
\end{equation}
where $\boldsymbol{M}$ is the TPR, %
$\boldsymbol{f} = (f_1,\dots,f_n)$ is a set of $n$ filler vectors and %
$\boldsymbol{r} = (r_1,\dots,r_n)$ is a set of $n$ role vectors. 
For each role-filler vector pair, which can be considered as an entity-relation pair, we \textit{bind} (or store) them into $\boldsymbol{M}$ by performing their outer product. 
Then, given an unbinding role vector $u_i$,
associated to the filler vector $f_i$, $f_i$ can be recovered by performing:
\begin{equation}
    \boldsymbol{M} u_i= \left[ \sum_{i} f_i\otimes r_i \right] u_i = \sum_{i} \alpha_{ij} f_i \propto f_i
\end{equation}
where $\alpha_{ij}\neq0$ if and only if $i=j$. It can be proven that the recovering is perfect if the role vectors are orthogonal to each other.
In our model, TPR-like \textit{binding}  and \textit{unbinding} methods are used to store and retrieve information from and to the TPR $\boldsymbol{M}$, which we will call memory.

\section{The StepGame Dataset}
To design a benchmark dataset that explicitly tests models' spatial reasoning ability and tackle the above mentioned problems, we build a new dataset named StepGame inspired by the spatial reasoning tasks in the bAbI dataset~\cite{weston2015towards}. %
The StepGame is a contextual QA dataset, where the system is required to interpret a story about several entities expressed in natural language and answer a question about the relative position of two of those entities. %
Although this reasoning task is trivial for humans, to equip current NLU models with such a spatial-ability remains still a challenge. 
Also, to increase the complexity of this dataset we model several form of \textit{distracting noises}. Such noises aim to make the task more difficult and force machine learning models that are trained on this dataset to be more robust in their inference process.  
\begin{figure}[!t]
  \centering
  \includegraphics[width=0.47\textwidth]{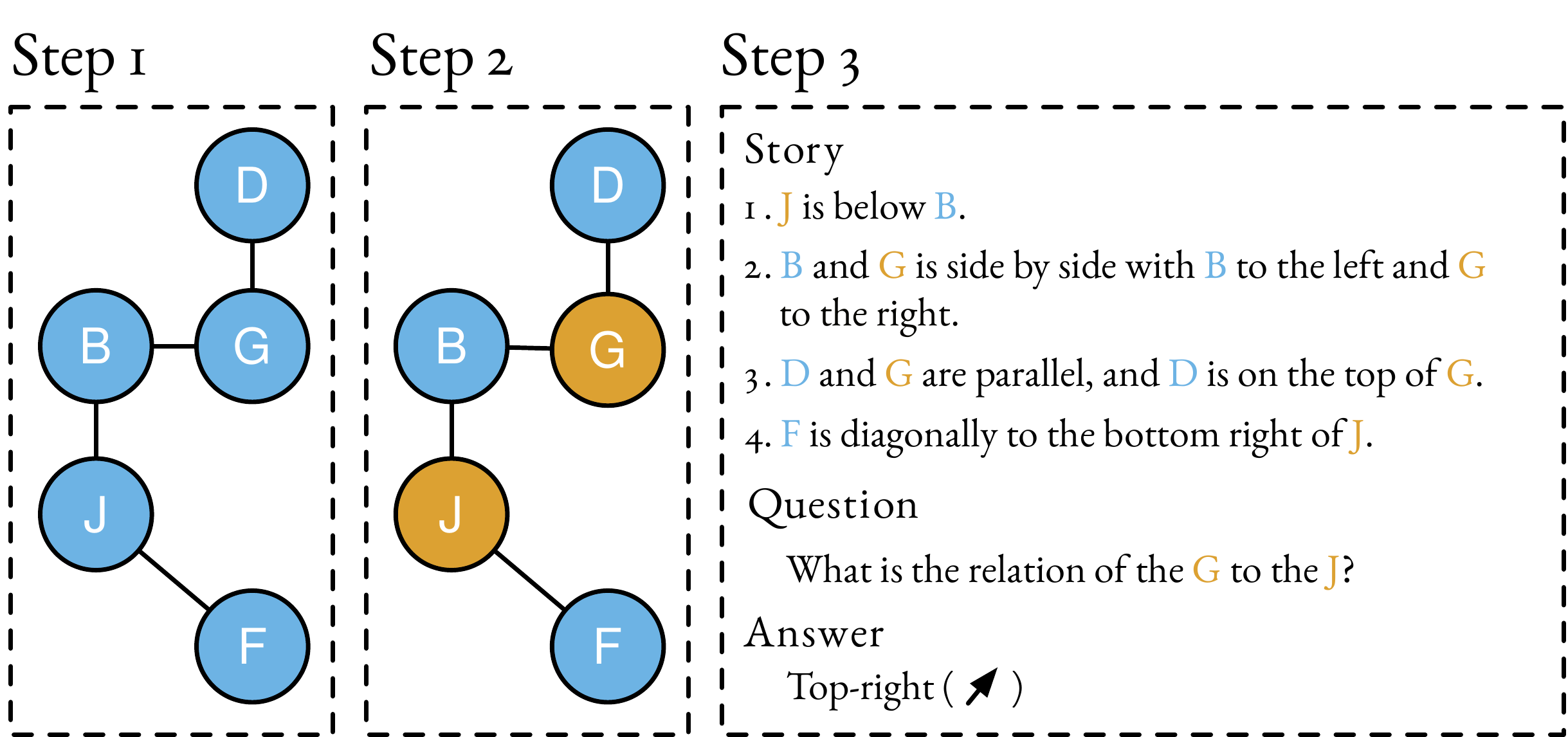}
  \caption{An example of the generation of a StepGame sample with $k=4$.}
  \label{generation_step}
\end{figure}

\subsection{Template Collection}

The aim of this crowdsourcing task is to find out all possible ways we can describe the positional relationship between two entities. 
The crowdworkers from Amazon Mechanical Turk were provided with an image visually describing the spatial relations of two entities and a request to describe these entities' relation. 
This crowdsourcing task was performed in multiple runs. 
In the first run, we provided crowdworkers with an image and two entities (e.g., A and B) and they were asked to describe their positional relation. 
From the data collected in this round, we then manually removed bad answers, and showed the remaining good ones as positive examples to crowdworkers in the next run. However, crowdworkers were instructed to avoid repeating them as an answer to our request. We repeated this process until no new templates could be collected. 
In total, after performing a manual generalization where templates discovered for a relation were translated to the other relations, we collected 23 templates for left and right relations, 27 templates for top and down relations, and 26 templates for top-left, top-right, down-left, and down-right relations.

\subsection{Data Generation}
\label{data_generation}

The task defined by the StepGame dataset is composed of several story-question pairs written in natural language. In its basic form, the story describes a set of $k$ spatial relations among $k+1$ entities, and it is structured as a list of $k$ sentences each talking about $2$ entities. The relations are $k$ and the entities $k+1$ because they define a chain-like shape. The question requests the relative position of two entities among the $k+1$ ones mentioned in the story. To each story-question pair an answer is associated. This answer can take $9$ possible values: \textit{top-left}, \textit{top-right}, \textit{top}, \textit{left}, \textit{overlap}, \textit{right}, \textit{down-left}, \textit{down-right}, and \textit{down}, each representing a relative position.
The number of edges between the two entities in the question ($\leq k$) determines the number of hops a model has to perform in order to get to the correct answer.

To generate a story, we follow three steps, as depicted in Figure~\ref{generation_step}. Given a value $k$ and a set of entities $\mathcal{E}$:

\paragraph{Step 1.} We generate a sequence of entities by sampling a set of $k+1$ unique entities from $\mathcal{E}$. 
Then, for each pair of entities in the sequence, $k$ spatial relations are sampled. These spatial relations can take any of the 8 possible values: top, down, left, right, top-left, top-right, down-left, and down-right. 
Because the sampling is unconstrained, entities can overlap with each other.
This step results in a sequence of linked entities that from now on we will call a chain. 

\paragraph{Step 2.} Two of the chain's entities are then selected at random to be used in the question.

\paragraph{Step 3.} From the chain generated in Step 1, 
we translate the $k$ relations into $k$ sentence descriptions in natural language. Each description is based on a randomly sampled crowdsourced template. We then shuffle these $k$ sentences to avoid potential distributional biases. These shuffled $k$ sentence descriptions is a called a story. From the entities selected in Step 2, we then generate a question also in natural language. Finally, using the chain and the selected entities, we infer the answer to each story-question pair.

\definecolor{wong_1}{rgb}{0.9019, 0.6235, 0}
\definecolor{wong_2}{rgb}{0, 0, 0}
\definecolor{wong_3}{rgb}{0.3372,0.7058,0.9137}
\definecolor{wong_4}{rgb}{0,0.6196,0.4509}

Given this generation process we can quickly calculate the complexity of the task before using the templates. This is possible because entities can overlap. 
Given $k$ relations, $k+1$ entities sampled from $\mathcal{E}$ in any order (${\color{wong_1}\bullet}$), %
8 possible relations between pairs of entities with 2 ways of describing them (${\color{wong_2}\bullet}$), e.g., A is on the left of B or B is on the right of A, %
a random order of the $k$ sentences in the story (${\color{wong_3}\bullet}$), and %
a question about 2 entities with 2 ways of describing it (${\color{wong_4}\bullet}$), 
the number of examples that we can generate is equal to: 
\begin{equation}\footnotesize
{\color{wong_1}(k+1)!\binom{|\mathcal{E}|}{k+1}} \cdot
{\color{wong_2}16^k}\cdot
{\color{wong_3}\frac{k!}{2}}\cdot
{\color{wong_4}2\binom{k+1}{2}}.
\end{equation}
The complexity of the dataset grows exponentially with $k$. The StepGame dataset uses $|\mathcal{E}| = 26$. For $k=1$ we have 10,400 possible samples, for $k=2$ we have more than 23 million samples, and so on. The sample complexity of the problem guarantees that when generating the dataset the probability of leaking samples from the training set to the test set diminishes with the increase of $k$. Please note that these calculations do not include templates. If we were to considering also the templates, the number of variations of the StepGame would be even larger.

\begin{figure}[!t]
  \centering
  \includegraphics[width=0.48\textwidth]{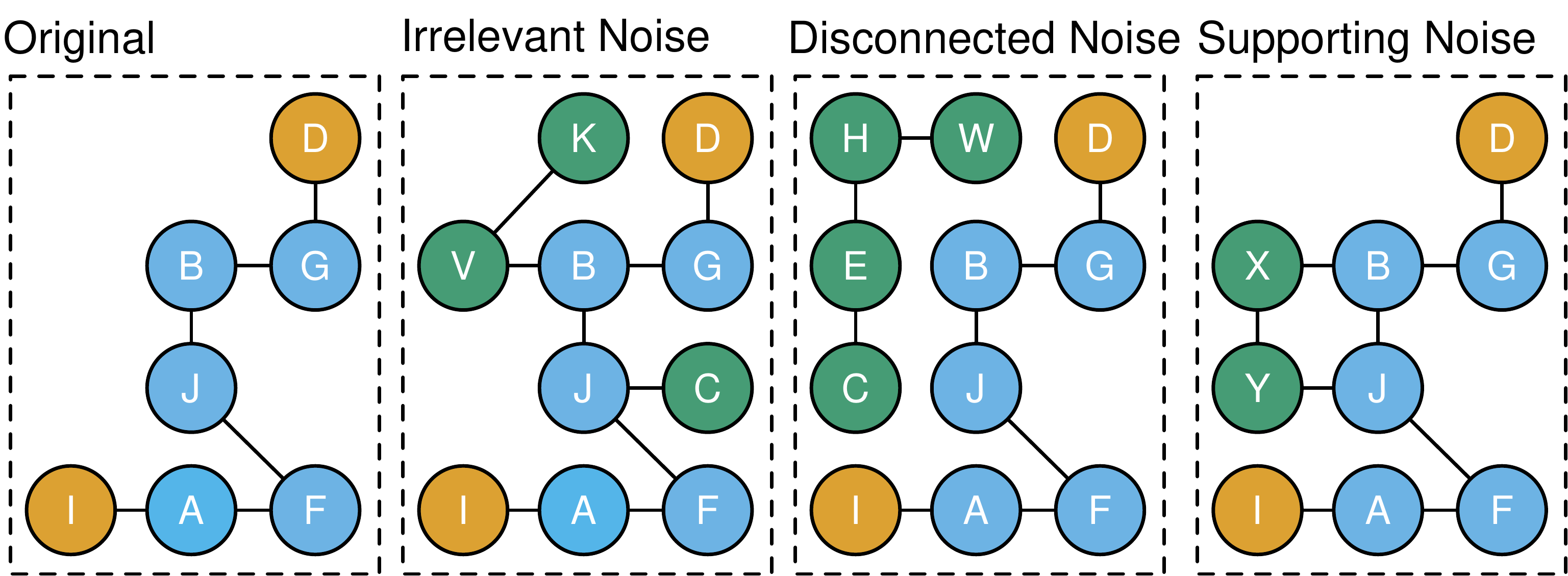}
  \caption{On the left-hand side we have the original chain. Orange entities are those targeted by the question. Beside, we show the same chain with the addition of noise. In green we represent irrelevant, disconnected and supporting entities.}
  \label{fig:example_noise}
\end{figure}

\subsection{Distracting Noise}
To make the StepGame more challenging we also include noisy examples in the test set.
We assume that when models trained on the non-noisy dataset make mistakes on the noisy test set, these models have failed to learn how to infer spatial relations. 
We generate three kinds of distracting noise: \textit{disconnected}, \textit{irrelevant}, and \textit{supporting}.
Examples of all kinds of noise are provided in Figure~\ref{fig:example_noise}. 
The irrelevant noise extends the original chain by branching it out with new entities and relations. 
The disconnected noise adds to the original chain a new independent chain with new entities and relations.
The supporting noise adds to the original chain new entities and relations that may provide alternative reasoning paths. We only add supporting noise into chains with more than 2 entities.
All kinds of noise have no impact on the correct answer.
The type and amount of noise added to each chain is randomly determined. The detailed statistics for each type of distracting noise are provided in the Appendix.

\section{The TP-MANN Model}
\label{sec:method}

\begin{figure*}[!t]
  \centering
  \includegraphics[width=0.75\textwidth]{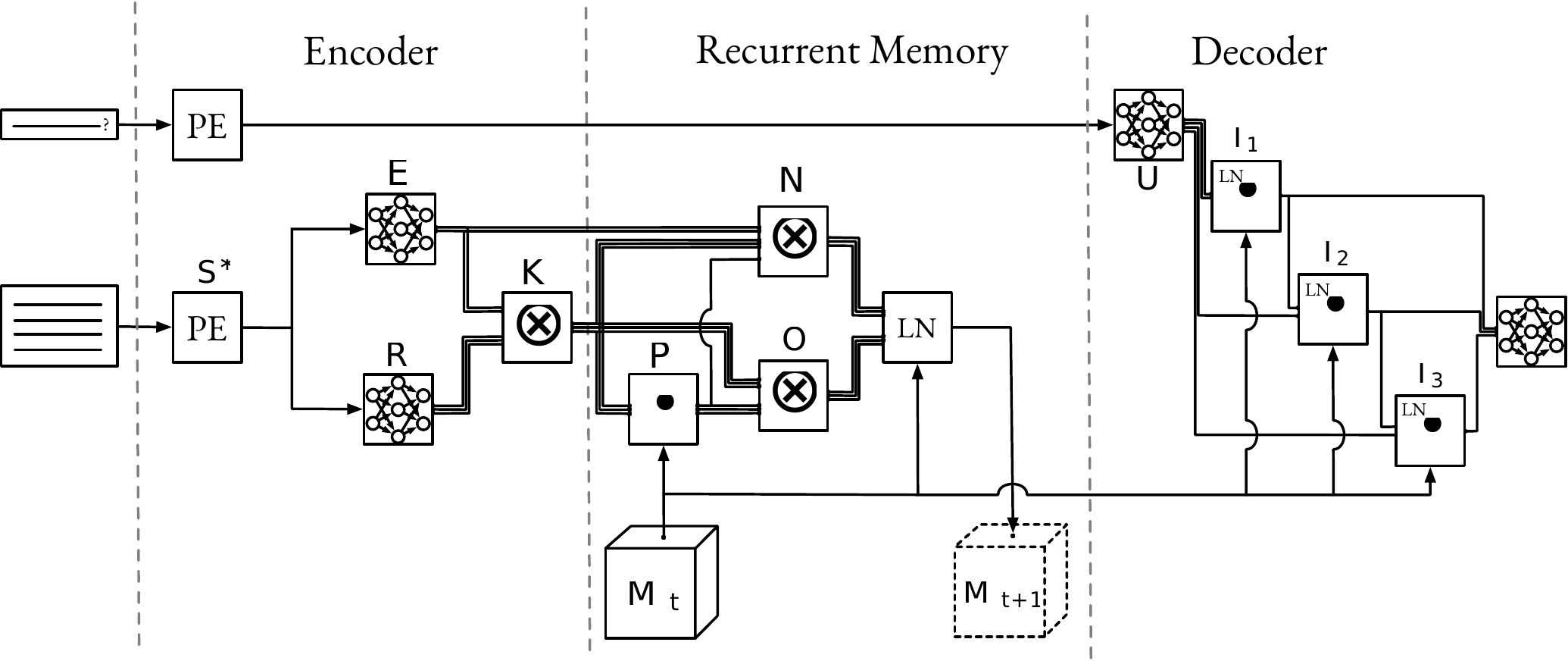}
  \caption{The TP-MANN architecture. PE stands for positional encoder, the sign in the box below the symbol $E$ represents a feed-forward neural network, the $\otimes$ sign represents the outer-product operator, the $\bullet$ sign represents the inner product operator, and LN represents a layer normalization. The $\otimes$, $\bullet$, and LN boxes implement the formulae as presented in Section \ref{sec:method}. Lines indicate the flow of information. Those without an arrow indicate which symbols are taken as input and are output by their box.}
  \label{fig:model}
\end{figure*}

In this section we introduce the proposed TP-MANN model, as shown in Figure~\ref{fig:model}. %
The model comprises three major components: a question and story encoder, a recurrent memory module, and a relation decoder. %
The encoder learns to represent entities and relations for each sentence in a story. 
The recurrent memory module learns to store entity-relation pair representations into the memory independently. It also updates the entity-relation pair representations based on the current memory and stores the inferred information. 
The decoder learns to represent the question and using the information stored in the memory recurrently infers the spatial relation of the two entities mentioned in the question.

It also has been shown that learned representations in the TPR-like memory could be orthogonal~\cite{schlag2018learning}. We use an example to illustrate the inspiration behind this architecture. A person may experience that when she goes back to her hometown and sees an old tree, her happy childhood memory about playing with her friends under that tree might be recalled. However, this memory may not be reminisced unless triggered by the old tree appearance.
In our model, unbinding vectors in the decoder module play the role of the old tree in the example, where unbinding vectors are learned based on the target questions. 
The decoder module unbinds relevant memories given a question via a recurrent mechanism. 
Moreover, although memories are stored separately, there are integration processes in  brains that 
retrieve information via a recursive mechanism. This allows episodes in memories to interact with each other 
\cite{kumaran2012generalization, schapiro2017complementary, koster2018big}. 
\paragraph{Encoder.}
The input of the encoder is a story and a question.
Given a input story $S=(s_1, \dots, s_m)$ with $m$ sentences and a question $q$ both described by words in a vocabulary $V$. Each sentence $s_i=(w_1, \dots, w_n)$ is mapped to learnable embeddings $(w^*_1, \dots, w^*_n)$. 
Then, a positional encoding (PE) is applied to each word embedding and then averaged together
$s^*_i=\frac{1}{n}\sum_{j=1}^{n} w^*_j \cdot p_j$, where $\{p_1,\dots,p_n\}$ are learnable position vectors, and $\cdot$ is the element-wise product. This operation defines $S^* \in \mathbb{R}^{m \times d}$, where each row of $S^*$ represents an encoded sentence and $d$ is the dimension of a word embedding. For the input question we convert it to a vector $q^* \in \mathbb{R}^{d}$ in the same way. 
For each sentence of the story in $S^*$, we learn entity and relation representations as:
\begin{align}
    E_i &= f_{e_i}(S^*),i=1,2, \\
    R_j &= f_{r_j}(S^*),j=1,2,3,
\end{align}
where $f_{e_i}$ are feed-forward neural networks that output entity representations $E_i \in \mathbb{R}^{m \times d_e}$ and 
$f_{r_j}$ are feed-forward neural networks that output relation representations $R_j \in \mathbb{R}^{m \times d_r}$. Finally, we define three search keys $K$ as:
\begin{align}
    K_1 &= E_1 \otimes R_1, \\
    K_2 &= E_1 \otimes R_2, \\
    K_3 &= E_2 \otimes R_3, 
\end{align}
where $K_1, K_2, K_3 \in \mathbb{R}^{m \times d_e \times d_r}$. Keys will be used to manipulate the memory in the next module and retrieve potential existing associations for each entity-relation pair.

\paragraph{Recurrent Memory Module.} %
To allow stored information to interact with each other, we use a recurrent architecture with $T$ recurrent-layers to update the TPR-like memory representation $\boldsymbol{M} \in \mathbb{R}^{d_e \times d_r \times d_e}$, where $\boldsymbol{M}$ contains trainable parameters. 
Through this recurrent architecture, existing episodes stored in memory can interact with new inferences to generate new episodes.
Different from many models like Transformer~\cite{vaswani2017attention} and graph-based models~\cite{kipf2016semi, velivckovic2017graph} where adding more layers in the model leads to a larger number of trainable parameters, our model will not increase the number of trainable parameters as the number of recurrent-layers increases.

At each layer $t$, given the keys $K$s, we extract pseudo-entities $P$s for each sentence in $S^*$. 
In the first layer ($t=0$), since there is no previous information existing in memory $\boldsymbol{M}_0$, the model just converts each sentence in $S^*$ as an episode and stores them in it ($\boldsymbol{M}_{1}$).
Then at the later layers ($t>0$), pseudo-entities $P$s build bridges between episodes in the current memory $\boldsymbol{M}_t$ and allow them to interact with potential entity-relation associations.
\begin{equation}
    P_{jt} = K_j \otimes M_t,  j=1,2,3,
\end{equation}
where $P_{jt} \in \mathbb{R}^{m \times d_e}$.
We then construct the memory episodes needed to be updated or removed. This is done after the first storage at $t=0$ so that all story information is already available in $\boldsymbol{M}$. 
These old episodes, $O_{jt} \in \mathbb{R}^{d_e \times d_r \times d_e}$, will be updated or removed to avoid memory conflicts that may occur when receiving new information:
\begin{equation}
    O_{jt} = K_j \otimes P_{jt},  j=1,2,3
\end{equation}
Afterwards, new episodes, $N_1, N_{2t}$ and $N_3 \in \mathbb{R}^{d_e \times d_r \times d_e}$, will be added into the memory:
\begin{align}
    N_1 = K_1 \otimes E_2, \\
    N_{2t} = K_2 \otimes P_{1t}, \\
    N_3 = K_3 \otimes E_1. 
\end{align}
Then we apply this change to the memory by removing (subtracting) old episodes and adding up the new ones to the now dated memory $\boldsymbol{M}_t$:
\begin{multline}
    \boldsymbol{M}_{t+1}=\text{LN}(\boldsymbol{M}_{t}+\\
    +N_1+N_{2t}+N_3-O_{1t}-O_{2t}-O_{3t}),
\end{multline}
where $LN$ is a layer normalization. 

\paragraph{Decoder.} %
The prediction is computed based on the constructed memory $\boldsymbol{M}$ at the last layer and a question vector $q$. To do this we follow the same procedure designed by \citet{schlag2018learning}:
\begin{align}
    U_j=f_j^{u}(q), j=1,2,3,4,
\end{align}
where $f_1^{u}$ is a feed-forward neural network that outputs a $d_e$-dimensional unbinding vector, and $f_2^{u},f_3^{u},f_4^{u}$ are feed-forward neural networks that output $d_r$-dimensional unbinding vectors. Then, the information stored in $\boldsymbol{M}$ will be retrieved in a recurrent way based on unbinding vectors learned from the question:
\begin{align}
    I_1=\text{LN}(\boldsymbol{M}_T\cdot U_1)\cdot U_2, \\
    I_2=\text{LN}(\boldsymbol{M}_T\cdot I_1)\cdot U_3, \\
    I_3=\text{LN}(\boldsymbol{M}_T\cdot I_2)\cdot U_4, \\
    \hat{v}=\text{softmax}(W_o \cdot \sum_{j=1}^{3} I_j).
\end{align}
A linear projection of trainable parameters $W_o \in \mathbb{R}^{\*|V| \times d_e}$ and a softmax function are used to map the extracted information into $\hat{v} \in \mathbb{R}^{\*|V|}$. Hence, the decoder module outputs a probability distribution over the terms of the vocabulary $\*V$.

\section{Experiments and Results}
In this section we aim to address the following research questions:
(\textbf{RQ1}) What is the degree of data leakage in the datasets?
(\textbf{RQ2}) How does our model behave with respect to state-of-the-art NLU models in spatial reasoning tasks? 
(\textbf{RQ3}) How do these models behave when tested on examples more challenging than those used for training? 
(\textbf{RQ4}) What is the effect of the number of recurrent-layers in the recurrent memory module?
Before answering these questions, we first present the material and baselines used in our experiments.
The software and data are available at: \url{https://github.com/ZhengxiangShi/StepGame}

\subsection{Material and Baselines}

In the following experiments we will use two datasets, the bAbI dataset and the StepGame dataset.
For the bAbI dataset we only focus on task 17 and task 19 and use the original train and test splits made of $10\,000$ samples for the training set and $1\,000$ for the validation and test sets.
For the StepGame dataset, we generate a training set made of samples varying $k$ from 1 to 5 at steps of 1, and a test set with $k$ varying from 1 to 10. Moreover, the test set will also contain distracting noise. The final dataset consists of, for each $k$ value, $10\,000$ training samples, $1\,000$ validation samples, and $10\,000$ test samples. 

We compare our model against five baselines: 
Recurrent Relational Networks (RRN)~\cite{palm2018recurrent}, %
Relational Network (RN)~\cite{santoro2017simple}, %
TPR-RNN~\cite{schlag2018learning}, %
Self-attentive Associative Memory (STM)~\cite{le2020self}, and 
Universal Transformer (UT)~\cite{dehghani2018universal}. 
Each model is trained and validated on each dataset independently following the hyper-parameter ranges and procedures provided in their original papers.
All training details, including those for our model, are reported in the Appendix.

\subsection{Training-Test Leakage}

To answer \textbf{RQ1} 
we have calculated the degree of data leakage present in bAbI and the StepGame datasets.
For the task 17, we counted how many samples in the test set appear also in the training set: 23.2\% of the test samples are also in the training set.
For task 19, for each sample we extracted the relevant sentences in the stories (i.e., those sentences necessary to answer the question correctly) and questions. Then we counted how many such pairs in the test set appear in the training set: 
80.2\% of the pairs overlap with pairs in the training set.
For the StepGame dataset, for each sample we extracted the sentences in the stories and questions. The sentences in the story are sorted in lexicographical order. Then we counted how many such pairs in the test set appear also in the training set before adding distracting noise and using the templates: 
1.09\% of the pairs overlap with triples in the training set. 
However, such overlap is all produced by the samples with $k=1$, which due to their limited number have a higher chance of being included in the test set.
If we remove those examples, the overlap between training and test sets drops to 0\%.

\subsection{Spatial Inference}
\label{sec:spatial_inference}

To answer \textbf{RQ2} and judge the spatial inference ability of our model and the baselines we train them on the bAbI and the StepGame datasets and compare them by measuring their test accuracy.

\begin{table}[!t]
\centering
\small
\begin{adjustbox}{max width=\textwidth}
\begin{tabular}{p{2.3cm}|cc|c}
\hline
\hline
\multicolumn{1}{c|}{} & Task 17 & Task 19 & Mean \\ \hline
RN    & 97.33$\pm$1.55 & 98.63$\pm$1.79 & 97.98 \\
RRN & 97.80$\pm$2.34 & 49.80$\pm$5.76 & 73.80 \\
STM & 97.80$\pm$1.06 & 99.98$\pm$0.05 & 98.89 \\
UT & 98.60$\pm$3.40 & 93.90$\pm$7.30 & 96.25 \\ 
TPR-RNN & 97.55$\pm$1.99 & 99.95$\pm$0.06 & 98.75 \\
\hline
Ours & \textbf{99.88$\pm$0.10} & \textbf{99.98$\pm$0.04} & \textbf{99.93} \\ \hline
\hline
\end{tabular}
\end{adjustbox}
\caption{Test accuracy on the task 17 and 19 of the bAbI dataset: Mean$\pm$Std over 5 runs.}
\label{tbl:babi_17_19}
\end{table}

In Table \ref{tbl:babi_17_19} we present the results of our model and the baselines on the task 17 and 19 of the bAbI dataset. 
The performance of our model is slightly better than the best baseline. However, due to the issues of the bAbI dataset, these results are not enough to firmly answer RQ2. 

\begin{table*}[!th]
\small
\centering
\begin{adjustbox}{max width=0.92\textwidth}
\begin{tabular}{l|rrrrl|r}
\hline
\hline
Model & \multicolumn{1}{c}{$k$=1} & \multicolumn{1}{c}{$k$=2} & \multicolumn{1}{c}{$k$=3} & \multicolumn{1}{c}{$k$=4} & \multicolumn{1}{c|}{$k$=5} & Mean \\ \hline
RN~\cite{santoro2017simple}          & 22.64$\pm$0.25 & 17.08$\pm$1.41 & 15.08$\pm$2.58 & 12.84$\pm$2.27 & 11.52$\pm$1.73 & 15.83 \\
RRN~\cite{palm2018recurrent}         & 24.05$\pm$4.48 & 19.98$\pm$4.68 & 16.03$\pm$2.89 & 13.22$\pm$2.51 & 12.31$\pm$2.16 & 17.12 \\
UT~\cite{dehghani2018universal}      & 45.11$\pm$4.16 & 28.36$\pm$4.50 & 17.41$\pm$2.18 & 14.07$\pm$2.87 & 13.45$\pm$1.35 & 23.68 \\
STM~\cite{le2020self}                & 53.42$\pm$3.73 & 35.96$\pm$4.45 & 23.03$\pm$1.83 & 18.45$\pm$1.87 & 15.14$\pm$1.56 & 29.20 \\
TPR-RNN~\cite{schlag2018learning}    & 70.29$\pm$3.03 & 46.03$\pm$2.24 & 36.14$\pm$2.66 & 26.82$\pm$2.64 & 24.77$\pm$2.75 & 40.81 \\ \hline
Ours                         & \textbf{85.77$\pm$3.18}  & \textbf{60.31$\pm$2.23} & \textbf{50.18$\pm$2.65} & \textbf{37.45$\pm$4.21} & \textbf{31.25$\pm$3.38} & \textbf{52.99} \\ \hline
\hline
\end{tabular}
\end{adjustbox}
\caption{Test accuracy on the StepGame dataset: Mean$\pm$Std over 5 runs.}
\label{tbls:step_game_hard}
\vspace{0.5em}
\end{table*}

\begin{table*}[!th]
\small
\centering
\begin{adjustbox}{max width=0.92\textwidth}
\begin{tabular}{l|rrrrl|r}
\hline
\hline
Model & \multicolumn{1}{c}{$k=6$} & \multicolumn{1}{c}{$k$=7} & \multicolumn{1}{c}{$k$=8} & \multicolumn{1}{c}{$k$=9} & \multicolumn{1}{c|}{$k$=10} & Mean \\ \hline
RN~\cite{santoro2017simple}       & 11.12$\pm$0.96 & 11.53$\pm$0.70 & 11.21$\pm$0.98 & 11.13$\pm$1.00 & 11.34$\pm$0.87 & 11.27 \\
RRN~\cite{palm2018recurrent}      & 11.62$\pm$0.80 & 11.40$\pm$0.76 & 11.83$\pm$0.75 & 11.22$\pm$0.86 & 11.69$\pm$1.40 & 11.56 \\
UT~\cite{dehghani2018universal}   & 12.73$\pm$2.37 & 12.11$\pm$1.52 & 11.40$\pm$0.92 & 11.41$\pm$0.96 & 11.74$\pm$1.07 & 11.88 \\
STM~\cite{le2020self}             & 13.80$\pm$1.95 & 12.63$\pm$1.69 & 11.54$\pm$1.61 & 11.30$\pm$1.13 & 11.77$\pm$0.93 & 12.21 \\
TPR-RNN~\cite{schlag2018learning} & 22.25$\pm$3.12 & 19.88$\pm$2.80 & 15.45$\pm$2.98 & 13.01$\pm$2.28 & 12.65$\pm$2.71 & 16.65 \\ \hline
Ours                              & \textbf{28.53$\pm$3.59}    & \textbf{26.45$\pm$2.95} & \textbf{23.67$\pm$2.78} & \textbf{22.52$\pm$2.36} & \textbf{21.46$\pm$1.72} & \textbf{24.53} \\ \hline
\hline
\end{tabular}
\end{adjustbox}
\caption{Test accuracy on StepGame for larger $k$s (only on the test set). Mean$\pm$Std over 5 runs.}
\label{tbl:step_game_hard_extra}
\end{table*}

In Table~\ref{tbls:step_game_hard} we present the results for the StepGame dataset. 
In this dataset, the training set is without noise but the test set is with distracting noise.
In the table we break down the performance of the trained models across $k$. In the last column we report the average performance across $k$.
Our model outperforms all the baseline models. %
Compared to Table~\ref{tbl:babi_17_19}, the decreased accuracy in Table~\ref{tbls:step_game_hard} demonstrates the difficulty of spatial reasoning with distracting noise. %
It is not surprising that the performance of all five baseline models decreases when $k$ increases, that is, when the number of required inference hops increases. 
We also report test accuracy on test sets without distracting noise in the Appendix.
\subsection{Systematic Generalization}

To answer \textbf{RQ3} 
we generate new StepGame test sets with $k \in \{6,7,8,9,10\}$ with distracting noise. 
We then test all the models jointly trained on the StepGame train set with $k \in \{1,2,3,4,5\}$ as in the Section~\ref{sec:spatial_inference}. We can consider this experiment as a zero-shot learning setting for larger $k$s. 

In Table~\ref{tbl:step_game_hard_extra} we present the performance of different models on this generalization task. Not surprisingly, the performance of all models degrades monotonically as we increase $k$. RN, RRN, UT and SAM fail to generalize to the test sets with higher $k$ values, while our model is more robust and outperforms the baseline models with a large margin. This demonstrates the better generalization ability of our model, which performs well on longer stories never seen during training. 

\subsection{Inference Analysis}

\begin{figure}[!t]
\centering
    \begin{subfigure}
        \centering
        \includegraphics[width=.407\linewidth]{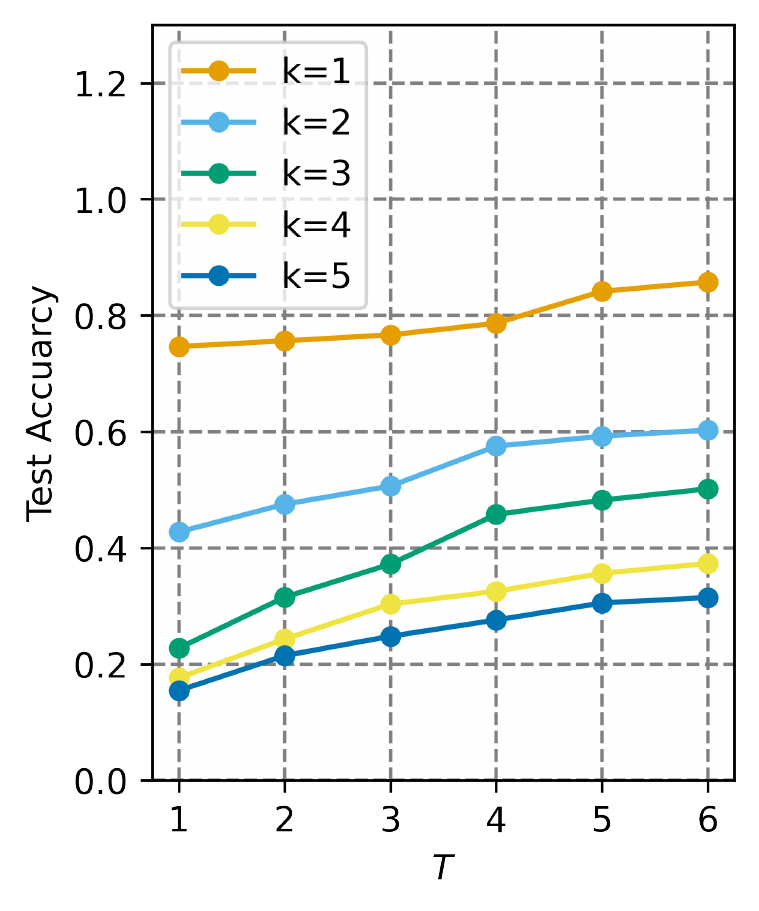}
        \label{fig:sub1}
    \end{subfigure}
    \begin{subfigure}
        \centering
        \includegraphics[width=.422\linewidth]{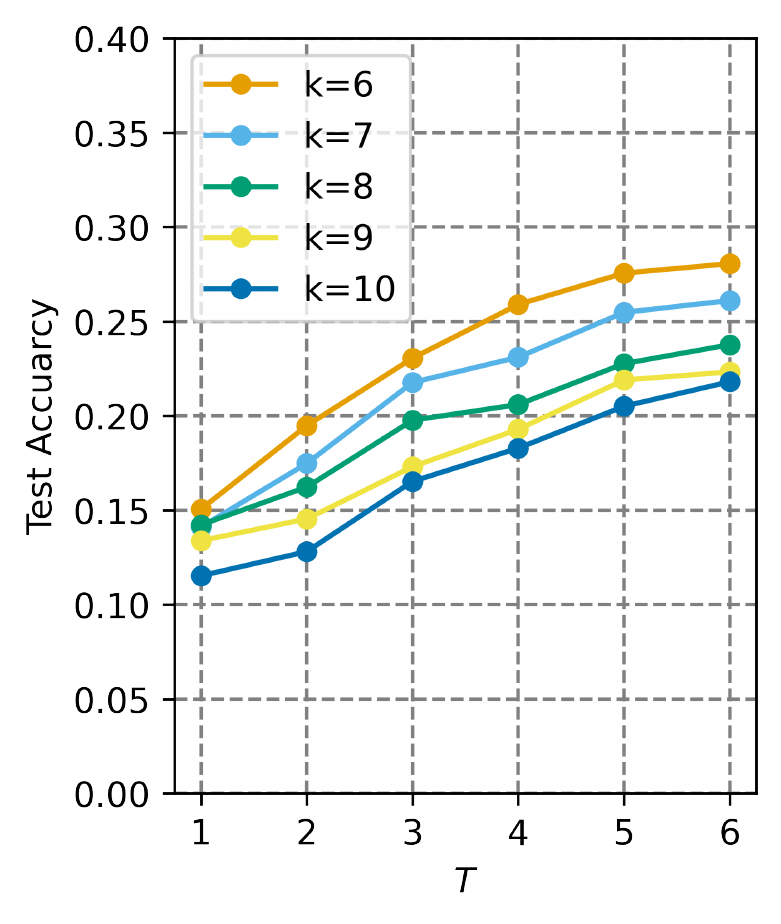}
        \label{fig:sub2}
    \end{subfigure}
\caption{Analysis of TP-MANN's number of recurrent-layers ($T$). The x-axis is $T$ with which the model has been trained. Each line represents a different value of $k$ of the StepGame dataset.}
\label{fig:inference_analysis}
\end{figure}

To answer \textbf{RQ4}, 
we conduct an analysis of the hyper-parameter $T$, the number of recurrent-layers in our model. 
We jointly train TP-MANN on the StepGame dataset with $k$ between 1 and 5 with number of $T$ between 1 and 6 and report the break down test accuracy for each value of $k$. 
These results are shown in the left-hand side figure of Figure~\ref{fig:inference_analysis}. 
The test sets with higher $k$ benefit more from a higher number of recurrent layers than those with lower $k$, indicating that recurrent layers are critical for multi-hop reasoning.
We also analyze how the recurrent layer structure affects systematic generalization. 
To do this we also test on a StepGame test set with $k$ between 6 and 10 with noise. These $k$s are larger than the largest $k$ used during training.
These results are shown in the right-hand side figure in Figure~\ref{fig:inference_analysis}.
Here we see that as $T$ increases, the performance of the model improves.
This analysis further corroborates that our recurrent structure supports multi-hop inference. 
It is worth noting, that the number of trainable parameters in our model remains unchanged as $T$ increases. 
Interestingly, we find that the number of recurrent-layers needed to solve the task is less than the length of the stories $k$ suggesting that the inference process may happen in parallel.

\section{Conclusion}
In this paper, we proposed a new dataset named StepGame that requires a robust multi-hop spatial reasoning ability to be solved and mitigates the issues observed in the bAbI dataset. 
Then, we introduced TP-MANN, a tensor product-based memory-augmented neural network architecture that achieves state-of-the-art performance on both datasets. 
Further analysis also demonstrated the importance 
of a recurrent memory module for multi-hop reasoning.

\newpage
\bibliography{ms}


\end{document}